%% file: main.tex
\documentclass[conference]{IEEEtran}
\usepackage{cite}
\usepackage{graphicx}
\usepackage{pgfplots}
\usepgfplotslibrary{groupplots}
\pgfplotsset{width=\linewidth,compat=1.9}
\usepackage{url}
\usepackage{tabularx, booktabs}
\usepackage{listings}
\usepackage{xcolor}
\usepackage{makecell} 						
\usepackage[acronym]{glossaries}
\usepackage[nointegrals]{wasysym}			
\usepackage[shortlabels]{enumitem}			
\usepackage{csquotes}
\usepackage{multirow}

\definecolor{light-gray}{gray}{0.95}
\lstdefinestyle{mystyle}{
    backgroundcolor=\color{light-gray},  
    basicstyle=\ttfamily\small
}
\definecolor{codegray}{rgb}{0.5,0.5,0.5}

\input{acronyms}

\begin{document}
\bstctlcite{IEEEexample:BSTcontrol} 

\title{Capability-Driven Skill Generation with LLMs: \\
A RAG-Based Approach for Reusing Existing \\
Libraries and Interfaces}


\author{
\IEEEauthorblockN{
    Luis Miguel Vieira da Silva\IEEEauthorrefmark{1},
    Aljosha Köcher\IEEEauthorrefmark{1},
    Nicolas König\IEEEauthorrefmark{1},
    Felix Gehlhoff\IEEEauthorrefmark{1}, 
    Alexander Fay\IEEEauthorrefmark{2}
}
\IEEEauthorblockA{
\IEEEauthorrefmark{1}Institute of Automation Technology\\
Helmut Schmidt University, Hamburg, Germany\\
Email: \{miguel.vieira, aljosha.koecher, felix.gehlhoff\}@hsu-hh.de\\}

\IEEEauthorblockA{
		\IEEEauthorrefmark{2} Chair of Automation \\
		Ruhr University, Bochum, Germany \\
		Email: alexander.fay@rub.de}
}

\maketitle

\begin{abstract}
Modern automation systems increasingly rely on modular architectures, with capabilities and skills as one solution approach. Capabilities define the functions of resources in a machine-readable form and skills provide the concrete implementations that realize those capabilities.
However, the development of a skill implementation conforming to a corresponding capability remains a time-consuming and challenging task. 
In this paper, we present a method that treats capabilities as contracts for skill implementations and leverages large language models to generate executable code based on natural language user input. 
A key feature of our approach is the integration of existing software libraries and interface technologies, enabling the generation of skill implementations across different target languages. 
We introduce a framework that allows users to incorporate their own libraries and resource interfaces into the code generation process through a retrieval-augmented generation architecture. 
The proposed method is evaluated using an autonomous mobile robot controlled via Python and ROS~2, demonstrating the feasibility and flexibility of the approach.
\end{abstract}


\section{Introduction}
\label{sec:introduction}
\input{sections/01_Introduction}

\section{Related Work} 
\label{sec:relatedWork}
\input{sections/02_RelatedWork}

\section{Method -- LLMCap2Skill}
\label{sec:method}
\input{sections/03_Method}

\section{Evaluation}
\label{sec:eval}
\input{sections/04_Evaluation}

\section{Conclusions and Future Work}
\label{sec:conclusion}
\input{sections/05_Conclusion}

\section*{Acknowledgment}
This research is funded by dtec.bw – Digitalization and Technology Research Center of the Bundeswehr as part of the project RIVA. dtec.bw is funded by the European Union – NextGenerationEU

\bibliographystyle{./bibliography/IEEEtran}
\bibliography{./bibliography/references} 

\end{document}

%% file: acronyms.tex
\newacronym{ABox}{ABox}{assertional box}
\newacronym{AAS}{AAS}{Asset Administration Shell}
\newacronym{llm}{LLM}{Large Language Model}
\newacronym{rag}{RAG}{Retrieval-Augmented Generation}
\newacronym{api}{API}{Application Programming Interface}
\newacronym{ros2}{ROS~2}{Robot Operating System 2}
\newacronym{iri}{IRI}{Internationalized Resource Identifier}
\newacronym{ODP}{ODP}{Ontology Design Patter}
\newacronym{OWL}{OWL}{Web Ontology Language}
\newacronym{SHACL}{SHACL}{Shapes Constraint Language}
\newacronym{TBox}{TBox}{terminological box}
\newacronym{UML}{UML}{Unified Modeling Language}

%% file: sections/01_Introduction.tex
Modern industrial automation systems increasingly require modularity and reconfigurability to flexibly adapt to changing tasks, environments and system configurations \cite{KHJ+_ReconfigurableManufacturingSystems_1999}. 
Whether in manufacturing or multi-robot scenarios, such systems typically consist of heterogeneous resources and components from various vendors, leading to considerable integration complexity. Each resource often exposes proprietary interfaces, protocols, and data formats, which must be harmonized to ensure reliable coordination and control.
A key requirement for seamless integration is the availability of machine-interpretable descriptions of each resource's functionality \cite{PGL+_Capabilitybasedsemanticinteroperabilityof_2019}. 

A common concept to realize this functionality description and implementation is the use of \emph{capabilities} and \emph{skills}. A capability provides an abstract, technology-independent description of a function a resource can offer (e.g., grasping an object), while a skill represents its concrete, executable implementation, exposed via a skill interface\cite{KBH+_AReferenceModelfor_15.09.2022b}. 
This distinction enables automated planning and orchestration based on abstract capabilities, while the execution is delegated to specific skills provided by the resources \cite{Cap_CSSModelExtensionsand_2025}.

Despite the potential of capabilities and skills, two major challenges remain for practical adoption. 
First, creating machine-interpretable capability models is complex and error-prone. 
Prior work has already demonstrated that structured engineering methods support the modeling of capabilities and mitigate these challenges (e.g., \cite{LAF+_TowardaMethodto_2024}). 
The second open challenge is that manually developing skills is time-consuming, error-prone, and requires deep technical knowledge about the target resource, its control interfaces and libraries. 
Even when a capability model is available, identifying appropriate control interfaces and implementing robust behaviors involves significant engineering effort.
This lack of automation slows down deployment and limits scalability in dynamic or modular environments where new functions must frequently be added or reconfigured.

Addressing this challenge requires methods for automatically generating skill implementations from formal capability models. Recently, \glspl{llm} have shown strong performance in generating code from natural language instructions, both in general application programming and for more specialized control languages \cite{FNS+_LLMBasedTestDrivenInteractiveCode_2024,KGA_ChatGPTforPLCDCS_2023}. 
Based on these findings, \glspl{llm} offer a promising solution to bridge the gap between capability specification and executable skills.
This paper presents an engineering approach that uses capability models and user-defined intents as input to automatically generate skill implementations. The generated skills integrate existing libraries and resource-specific interfaces. To break down this objective, the following research questions are addressed:

\begin{enumerate}
[start=1,leftmargin=*,label=\textbf{RQ \arabic*:}]
    \item How can capability models be used to generate skill skeletons that reflect the capability constraints?
    \item How can executable skill behavior be synthesized from user intent and capability context using \glspl{llm}?
    \item How can existing system interfaces and libraries be leveraged during code generation?
\end{enumerate}

The remainder of this paper begins with an overview of related work in Section~\ref{sec:relatedWork}. Section~\ref{sec:method} introduces the \textbf{LLMCap2Skill} method in detail. 
Section~\ref{sec:eval} presents the evaluation setup and results. 
The paper concludes with a discussion and outlook in Section~\ref{sec:conclusion}.

%% file: sections/02_RelatedWork.tex
While research on capabilities and skills has a history of almost three decades, initial efforts to standardize the relevant terms and create a unified model were only presented in \cite{KBH+_AReferenceModelfor_15.09.2022b}. 
In this work, a reference model defined by a working group is proposed, which defines the core elements of capabilities and skills and represents them using an abstract UML model. This foundational model addresses the previously missing lack of consistency and provides a conceptual framework for further developments.
In \cite{KHV+_AFormalCapabilityand_9820209112020}, we initially introduced an OWL ontology to model capabilities and skills. This ontology has been continuously extended and conforms with the meta model introduced in \cite{KBH+_AReferenceModelfor_15.09.2022b}. 
The current version of the ontology is structured into three layers: At the most abstract layer, the \emph{CSS ontology} offers a one-to-one implementation of the reference model in \cite{KBH+_AReferenceModelfor_15.09.2022b}.
The \emph{CaSk ontology} is an extension of CSS, which extends abstract concepts (e.g., \texttt{CapabilityConstraint} and \texttt{StateMachine}) by using separate standard-backed ontologies.
For this paper, the CaSk ontology acts as the semantic foundation to represent capabilities.

Even with a reference model, one issue with capabilities and skills still is the additional effort to develop resource functions as skills and create the machine-interpretable models \cite{Cap_CSSModelExtensionsand_2025}. In \cite{Cap_CSSModelExtensionsand_2025}, abstract methods to create capabilities are described, but no concrete solutions to generate capability models or skill implementations are given.
For such model or code generation tasks, \glspl{llm} have proven highly promising \cite{FNS+_LLMBasedTestDrivenInteractiveCode_2024}.

We presented an \gls{llm}-based engineering method to generate capabilities from natural-language task descriptions in \cite{LAF+_TowardaMethodto_2024}. 
However, this method does not generate skill implementations. 
The present paper pursues a different goal: generating skill implementations based on given capability descriptions.

In recent years, a variety of works has emerged that utilize \glspl{llm} for code generation, and which can be extended for skills.
In \cite{CRS+_LLMGeneratedMicroserviceImplementationsfrom_13.02.2025}, Chauhan et al. use an OpenAPI specification as a blueprint to derive endpoints, schemas, routes, and validation directly from it. The approach in \cite{CRS+_LLMGeneratedMicroserviceImplementationsfrom_13.02.2025} is designed for general IT requirements and doesn’t cover industrial-automation needs, e.g., directly interfacing with hardware. Also, it cannot make use of existing libraries unknown to the \gls{llm}.

The study in \cite{KGA_ChatGPTforPLCDCS_2023} was one of the first to prove that \glspl{llm} can also handle the specialized domain of automated control \cite{KGA_ChatGPTforPLCDCS_2023}. In this study, ChatGPT was used to generate control code for 100 tasks. The results indicate that \glspl{llm} can produce correct and executable code, sometimes even adding relevant contextual knowledge. However, the generated solutions require careful validation to ensure correctness \cite{KGA_ChatGPTforPLCDCS_2023}.

While the approach in \cite{KGA_ChatGPTforPLCDCS_2023} relies on rather simple prompting to generate control code with ChatGPT, LLM4PLC is a more advanced pipeline that combines prompt engineering, model fine-tuning, and automated verification. This results in a higher code quality and ensures compliance with the strict correctness and safety requirements of industrial applications \cite{FDM+_LLM4PLC:HarnessingLargeLanguage_2024}.

More recently, Agents4PLC \cite{LZW+_Agents4PLC:AutomatingClosedloopPLC_2024} introduces a multi-agent framework that fully automates control code generation and verification. Unlike LLM4PLC, which focuses on design-level checks, Agents4PLC performs formal code-level verification in a closed-loop workflow, improving autonomy and reliability.

In \cite{KGH+_LLMbasedandRetrievalAugmentedControl_2024}, Koziolek et al. investigate how control code can be generated using \glspl{llm} with consideration of existing libraries. Since these libraries are not inherently known to the \gls{llm}, Koziolek et al. employ a \gls{rag} approach. 
\gls{rag} combines \glspl{llm} with a retrieval component that searches a vector database of embedded information. At inference time, the user query is embedded, relevant documents are retrieved based on similarity, and the language model generates its output conditioned on both the query and the retrieved context. \gls{rag} enhances factual accuracy, allows access to up-to-date or domain-specific knowledge, and reduces hallucinations in the generated output.
The approach in \cite{KGH+_LLMbasedandRetrievalAugmentedControl_2024} focuses solely on generating control code targeting the IEC 61131 standard and its only input is a user task, i.e., it does not make use of an existing specification such as in \cite{CRS+_LLMGeneratedMicroserviceImplementationsfrom_13.02.2025}. In this paper, we present a similar approach to \cite{KGH+_LLMbasedandRetrievalAugmentedControl_2024} that conforms to a capability description as a contract and enables code generation across multiple target languages and supporting different libraries and existing interfaces. This is achieved by explicitly incorporating a step to embed library or interface information into our method.

While there are promising \gls{llm}-based code-generation approaches, many fall short for industrial automation: most ignore existing model-based specifications; some lack support for hardware interfaces or libraries; and others are hard-coded to a single language or framework, lacking extensibility.

%% file: sections/03_Method.tex
The manual development of skill implementations requires in-depth knowledge of the hardware in use and its corresponding control interfaces. 
In addition, it demands a detailed understanding of the structural and behavioral aspects of a skill itself -- such as its state machine logic, skill interface for interaction, and ontological representation.
To automate this challenging and time-consuming task, we present \textbf{LLMCap2Skill}, our method for generating skills. 
It consists of three main steps: 
(1) user input,
(2) generation of an API documentation, and
(3) automated \gls{rag}-based skill generation using an \gls{llm}.

The first essential input is the capability ontology, which provides an abstract description of a function by specifying its inputs and outputs as products or information, along with their properties. 
This capability description serves as the structural basis for the implementation of the skill -- the executable counterpart of the abstract capability. 
However, the capability alone is not sufficient for code generation, as it does not include any information about the concrete behavior required for execution.
Therefore, an additional, simple natural language skill specification is required as user input in step~1, detailing the intended behavior of the skill to enable meaningful implementation.

While the capability description defines the structural frame and the skill specification provides the intended behavior, it still remains unclear how to interact with the resource that provides both the capability and the final skill.
To enable actual control, additional information about the resource and its interfaces is required. 
Therefore, in step~2, the resource interfaces are automatically identified and structured into an API document to be used in the subsequent code generation.

In step~3, the actual skill implementation is generated using \gls{rag}. 
To do this, the method automatically searches the previously generated API document for resource interfaces that best match the given capability.
These relevant resource interface descriptions, together with the capability ontology and the skill specification, are passed to the \gls{llm}. 
In addition, we leverage the \emph{pySkillUp} framework developed in our previous work~\cite{VKT+_APythonFrameworkfor_2023}.
This framework introduces annotations for specifying skill metadata, parameters, and states, enabling the automatic generation of the ontology, skill interface, and state machine. 
Together, these prompt inputs allow the \gls{llm} to generate a complete and valid skill implementation that conforms to the defined behavior and technical context.
The algorithms that implement the method and results of the evaluation are available online\footnote{https://github.com/CaSkade-Automation/Cap2Skill}.

To demonstrate the method, we focus on a representative implementation using the \gls{ros2} framework and the Python programming language.
This combination is very common in autonomous mobile robots and provides a practical foundation for demonstrating the method in a realistic system context. 
The following subsections explain each step of the method in detail, concluding with an outline of its extension to other frameworks and programming languages.

\subsection{User Input}
\label{subsec:userinput}
To generate an executable skill from an abstract capability description, additional user input is required in step~1.
The required user inputs include:
\begin{itemize}
    \item the capability ontology itself,
    \item a skill specification describing the intended behavior,
    \item the programming language: Python, and
    \item the name of the target implementation framework: \gls{ros2}.
\end{itemize}
The skill specification is essential, as the capability does not contain any information about the expected behavior.
Only in very simple cases -- such as \emph{get\_position} -- can the behavior be directly inferred from the capability name and structure by the \gls{llm}.
For most abstract capabilities, however, an explicit skill specification is necessary, as a skill may require multiple control actions that cannot be derived from the capability description alone.
The skill specification consists of two parts, which are entered through a structured input form designed to simplify the process: 
First, metadata such as the skill interface type through which the skill should be invoked (e.g., REST), which can be selected from a drop-down menu listing all supported types. Additionally, users may provide an optional natural language description of the skill to improve clarity and documentation, though it is not required for skill generation. 
Second, the form allows users to define a behavioral definition of the skill logic within each state of the skill’s state machine, which follows the standardized \emph{PackML} state machine as defined in the CaSk ontology. 
For each state, a text input can be used to describe the expected behavior in natural language. 
While behavior can be defined for all states, specifying the behavior for the \emph{execute} state is mandatory, as it represents the core functionality. 
The level of detail and clarity in this natural language input has a significant impact on the quality of the generated skill, with more precise and specific formulations leading to more accurate and robust results.

Furthermore, the programming language Python and target framework \gls{ros2} determine the automated discovery of available \gls{ros2} interfaces in step~2 and subsequent implementation of executable skills with \gls{ros2} in Python in step~3. 
All user inputs are used in the final prompt to guide the \gls{llm} in generating an appropriate skill implementation. 

\subsection{API Document Generation and Preparation}
\label{subsec:api}
To generate a skill implementation for a given capability, knowledge of how to control the underlying resource is essential. 
Therefore, the resource interfaces -- in this case, the available \gls{ros2} interfaces -- are automatically queried.
The resulting resource interfaces are then automatically structured into a consistent API documentation used for the next step. 

To retrieve all available \gls{ros2} interfaces, we developed a script that launches a dedicated \gls{ros2} node. 
This node automatically queries the resource for all active topics, services, and actions, and compiles the results into a structured JSON document.
For each interface, the script records its name and message type, along with detailed information about the corresponding message definition -- including all relevant parameters required to interact with the interface.
In order to use this script, the node must be executed within the same \gls{ros2} network as the resource to be analyzed, as it relies on runtime introspection to access the resource's available interfaces. 
The extracted information is automatically stored in objects and organized into a dictionary of resource interfaces. 
This step is performed once during system integration and is not required at runtime. 
The resource interface report is persistent and can be reused across multiple skill generations for the same resource.

As a preparation for the subsequent \gls{rag} step, we use an \gls{llm} to generate uniformly structured descriptions of all collected resource interfaces. 
The original resource interface representations vary significantly in terms of length, level of detail, and formatting -- ranging from brief type definitions to complex multi-level message structures.
This heterogeneity makes direct similarity comparison with capability descriptions unreliable for identifying relevant interfaces required to implement the skill.
To address this, each resource interface is enriched with a compact, consistently structured natural language description generated by an \gls{llm}.
The prompt used for generating resource interface descriptions instructs the \gls{llm} to include the following elements:
(1) the resource module to which the interface relates, (2) type of tasks for which the interface is relevant, and (3) typical entities that use or interact with the interface.  
These descriptions are later used as input for the retrieval process in step~3.

Additionally, we introduce an optional automated resource interface relevance check to determine whether a given interface is actually involved in controlling the resource. 
The relevance check is also performed by an \gls{llm}.
Resource interfaces are considered irrelevant if they are used exclusively for debugging, logging, introspection or metadata exchange. 
In contrast, resource interfaces are deemed relevant if they actively participate in control processes, receive control-related messages, or are part of known control mechanisms.
If a resource interface is classified as irrelevant, it is excluded from the \gls{rag} step, reducing the size of the search space and improving efficiency. 
In the following step, both the full interface specifications and their natural language descriptions are used for skill generation.

\subsection{Skill Generation with Retrieval-Augmented Generation}
\label{subsec:RAG}

\begin{figure*}[htb]
    \centering
    \includegraphics[width=\linewidth]{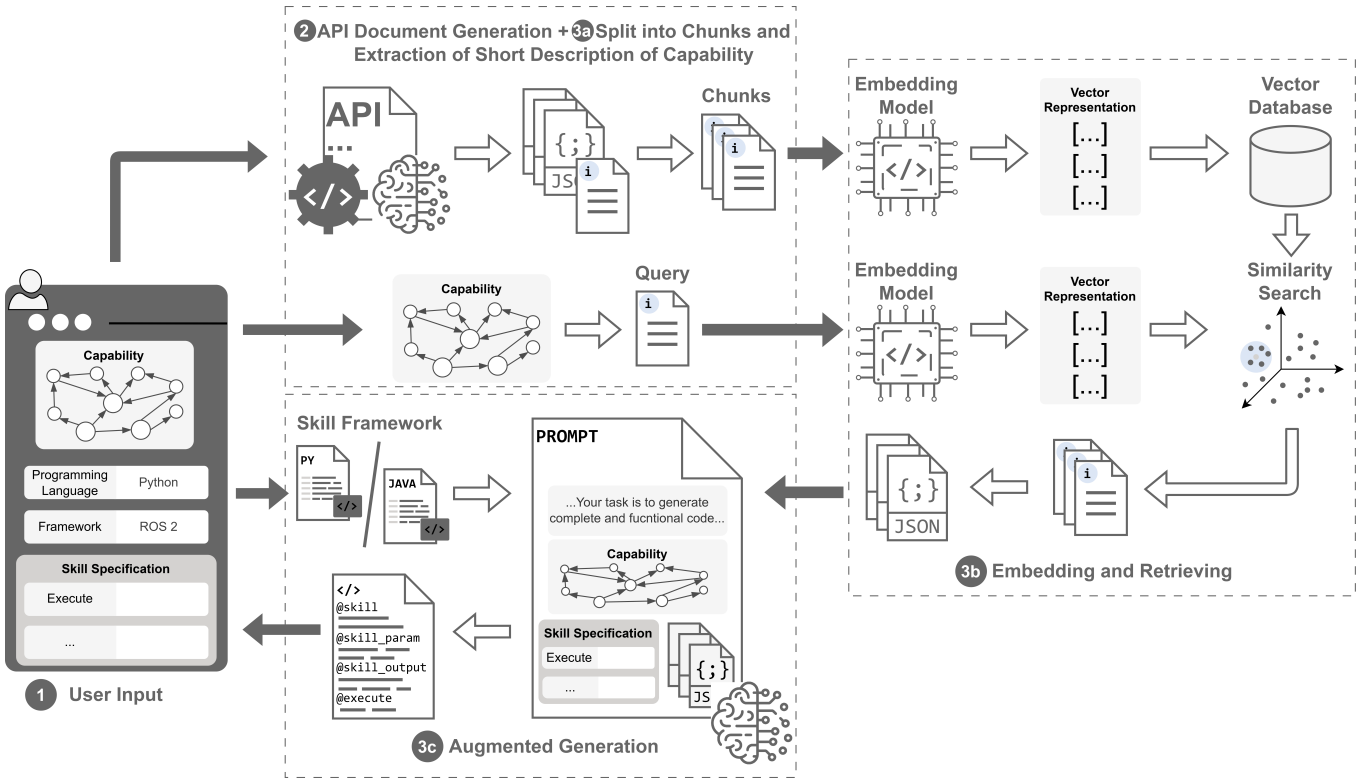}
    \caption{Skill generation using \gls{rag} by retrieving relevant resource interfaces and combining them with user input to construct a prompt for the \gls{llm}.}
    \label{fig:rag}
\end{figure*}

In the final step of the proposed method, the actual skill implementation is generated as shown in Figure~\ref{fig:rag}.
To identify the most relevant resource interfaces for the capability in question, the previously generated API documentation is processed using a \gls{rag} approach. 
The capability description, skill specification, and the retrieved resource interfaces are then combined into a structured prompt, which is submitted to an \gls{llm} to implement the final skill.

In step 3a, from the generated API documentation, only the uniform textual descriptions of resource interfaces -- previously generated by the \gls{llm} -- are extracted and processed individually as input chunks for embedding.
Similarly, from the capability ontology, only \texttt{rdfs:comment}, i.e., a natural-language description, is extracted as input for the embedding model. 
Both resource interface descriptions and capability definitions are reduced to natural language format before embedding.

In step 3b, the embedding model transforms each input text into a compact vector that captures its semantic meaning. 
These vectors are stored in a vector database, which enables fast similarity searches in a high-dimensional vector space. 
Subsequently, similarity search is performed to identify the resource interfaces most relevant to the given capability.
Resource interface and capability descriptions are compared in the embedding space, where semantic closeness is measured using distance metrics such as cosine similarity.
This process narrows the complete set of resource interfaces down to a few candidates that are most likely relevant for implementing the given capability. 
The full resource interface documentations are selected based on the retrieved descriptions to provide the \gls{llm} with the necessary information to invoke and interact with the interfaces in the subsequent prompt used for skill generation. 
The embedding and retrieval functionality was implemented using the open-source library \emph{LangChain}\footnote{https://python.langchain.com/docs/introduction/}.

Finally, in step 3c, the resulting set of relevant resource interfaces is combined with the complete user input -- including the capability, programming language, framework, and skill specification -- and used to construct a prompt for the \gls{llm}.
The prompt also includes default instructions on how to structure the skill implementation. 
To further support skill automation regarding ontology, skill interface, and state machine generation, the prompt is extended with additional framework-specific guidance by including an explanation of the \emph{pySkillUp} framework.
As a result, the \gls{llm} can generate a complete and valid skill that conforms to the framework specification.
Additionally, the prompt follows a few-shot prompting technique by including three examples. 
Few-shot prompting refers to providing the \gls{llm} with a small number of input–output examples to guide the generation process.
Each example contains a capability, a skill specification, and the resulting implementation. 
The prompt is finally submitted to the \gls{llm}, and the generated skill code is returned to the user. 

\subsection{Method Generalizability}
\begin{figure}[htb]
    \centering
    \includegraphics[width=\linewidth]{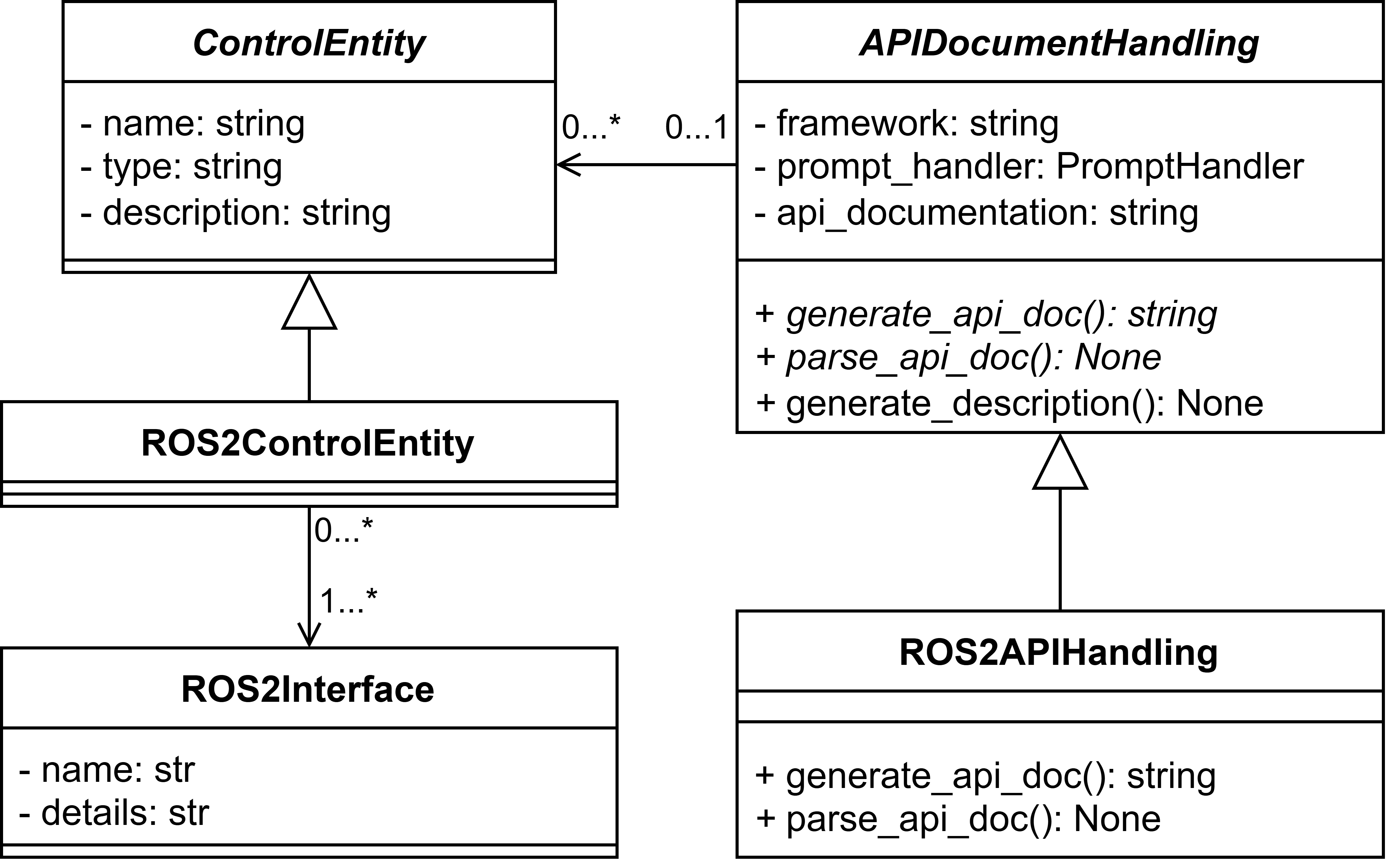}
    \caption{Class structure for resource interface documentation templates and \gls{ros2} specific implementations.}
    \label{fig:interface_classes}
\end{figure}

While the method has been exemplified for Python and \gls{ros2}, it is designed to be extensible and adaptable to other programming languages and frameworks. 
The user input in step~1 includes the specification of the target programming language and framework, which determines how the skill will be implemented and how resource interfaces are discovered and documented.
Resource interfaces can either be provided directly in step~1 -- for example, as a structured document or a PDF -- or derived in step~2 using custom extraction routines.
If the resource exposes its interfaces at runtime -- for example via introspection mechanisms or standardized service registries -- this information can be automatically queried  within the resource's network environment and generate a structured API documentation. 
Therefore, we provide a modular and extensible architecture to support alternative frameworks. 
Specifically, a set of abstract base classes as templates to support the documentation process, as illustrated in Figure~\ref{fig:interface_classes}. 
The \texttt{ControlEntity} class defines the structure of an individual control interface and can be extended for framework-specific cases (e.g., \texttt{ROS2ControlEntity} for \gls{ros2}). 
Given the complexity of interface structures in \gls{ros2} -- where communication with an entity is implemented via specific message types -- we further define the \texttt{ROS2Interface} class, which stores details such as message parameter types.
The second main class, \texttt{APIDocumentHandling}, is responsible for managing all resource interfaces and interacting with the \gls{llm}. 
When using these templates, developers must implement the specific querying and storage logic with \texttt{generate\_api\_doc} and \texttt{parse\_api\_doc} methods, as demonstrated in the \texttt{ROS2APIHandling} implementation. 
Our \gls{ros2} implementation extracts all available topics, services, and actions from the running \gls{ros2} resource in method \texttt{generate\_api\_doc}, parses them into structured interface objects and stores them in the dictionary-based report in \texttt{parse\_api\_doc}.

In the final step of the method, skill generation is adapted based on the selected programming language.
Where available, an appropriate skill framework is included in the prompt to support automatic generation of skill interface, ontology, and state machine.
For example, \emph{pySkillUp} is used for Python, and \emph{skillUp} is available for Java.
In these cases, the prompt includes a short explanation of the respective framework, as described in Subsection~\ref{subsec:RAG}.
For other languages without a predefined skill framework, the user may provide their own skill framework structure.
If no skill framework is used, only the core behavior of the skill is generated, which can then be manually integrated into the desired execution environment.

%% file: sections/04_Evaluation.tex
In this section, we evaluate the proposed method \textbf{LLMCap2Skill} with regard to its ability to generate executable and functionally correct skills for a mobile robot programmed using \gls{ros2}.

\subsection{Setup and Procedure}
To systematically assess the effectiveness of \textbf{LLMCap2Skill}, an evaluation framework was designed that covers both technical correctness and functional performance of the generated skills. 
The evaluation setup includes the following core components, which are described in more detail in the subsequent paragraphs: 
\begin{itemize}
    \item \textbf{Platform:} Mobile robot (Neobotix MMO-700), simulated in Gazebo with \gls{ros2}.
    \item \textbf{Capabilities:} Nine capabilities modeled using the CaSk ontology in Turtle syntax, which are listed in Table~\ref{tab:capabilityOverview}. 
    \item \textbf{Target language and framework:} Python and \gls{ros2}.
    \item \textbf{LLMs:} \emph{gpt-4o} and \emph{o3-mini}.
    \item \textbf{Prompting method:} few-shot prompt (three examples).
    \item \textbf{Embeddings:} \emph{text-embedding-3-large} stored in a \emph{Chroma} vector store.
    \item \textbf{Similarity search:} Cosine similarity (top four results).
\end{itemize}

In the first step of the proposed method, each of the nine capabilities of varying complexity was modeled, providing the expected inputs, outputs and constraints for the corresponding skill. 
Three of nine capabilities -- \emph{get-position}, \emph{set-velocity}, and \emph{navigate-to-point} -- serve as few-shot examples within the prompt for skill generation. 
They represent different \gls{ros2} interaction patterns: (1) \emph{get-position} requires subscription to a topic, (2) \emph{set-velocity} requires both publishing and subscribing, as velocity commands are published to a control topic, while feedback is received via subscription to verify the applied velocity, (3) \emph{navigate-to-point} is the most complex, requiring the use of an action client and handling various feedback and result messages.
The remaining six capabilities target various aspects of robot operation, such as navigation, mapping, and manipulation, and were used as input for skill generation. 

\begin{table}[htb]
    \caption{Overview of the capabilities used}
    \label{tab:capabilityOverview}
    \renewcommand*{\arraystretch}{1.2}
        \begin{tabularx}{\linewidth}{>{\raggedright\arraybackslash}p{2cm} p{3cm} p{1cm} p{1cm}}
    \toprule
    Name & Description & Inputs & Outputs \\
    \midrule
    E1: Set Velocity & Set robot's velocity in forward direction & $vel_{in}$ & $vel_{out}$ \\
    E2: Get Position & Retrieve robot's position & - & $pos_{out}$ \\
    E3: Navigate to Point & Navigate robot to a desired goal point & $pos_{in}$ & $pos_{out}$ \\
    \midrule
    C1: Move Forward & Set robot's velocity based on desired distance and travel time & \makecell{$dist_{in}$ \\  $time_{in}$} & \makecell{$dist_{out}$ \\ $time_{out}$ \\ $vel_{out}$} \\ 
    C2: Get Object & Retrieve distance and orientation to robot's nearest object & - & \makecell{$obs_{dist}$ \\ $obs_{degree}$} \\
    C3: Rotate & Set robot's angular velocity until desired orientation is reached & \makecell{$degree_{in}$ \\ $vel_{in}$} & $degree_{out}$ \\
    C4: Collision Avoidance & Move robot with specific motion pattern for a set time and stop if obstacle distance falls below min & \makecell{$vel_{in}$ \\ $min_{dist}$ \\  $time_{in}$} & \makecell{$vel_{out}$ \\ $obs_{dist}$ \\ $obs_{degree}$ \\ $time_{out}$} \\
    C5: Mapping & Map desired area by moving the robot in that area & $area_{in}$ & $map_{out}$ \\
    C6: Move to Point & Move manipulator to a desired goal point & $pos_{in}$ & $pos_{out}$ \\
    \bottomrule
    \end{tabularx}
\end{table}

For each capability, a corresponding skill specification was defined, including a description of the intended behavior within the states of the state machine.      
For instance, in the case of the \emph{move-forward} skill, the behavior of the execute state was specified as follows:

\begin{quote} \emph{Set the velocity of the robot to a calculated value for the desired time only and then reset it. If the calculated velocity exceeds the maximum velocity, set the maximum velocity and calculate the new time necessary to travel the desired distance.} \end{quote}

As part of the second step in our method to provide technical control information, the API documentation for the robot was generated once with \gls{ros2}. 
During the documentation phase, approximately 60 resource interfaces were excluded due to being irrelevant for control-related tasks. 
For the remaining $\sim130$ resource interfaces (e.g., \texttt{cmd\_vel} or \texttt{odom}), natural language descriptions were generated using \emph{gpt-4o}. 
 
Based on this documentation, the capabilities and the skill specifications, as well as the user-defined target programming language Python and framework \gls{ros2}, six skill implementations were generated in the last step of our method using two different \glspl{llm}: \emph{gpt-4o} and \emph{o3-mini}, both leveraging the \gls{rag} mechanism and few-shot prompting.
To ensure deterministic behavior, \emph{gpt-4o} was configured with a \emph{temperature} of 0 and \emph{top\_p} set to 1. 
In the future, the underlying \gls{llm} can be flexibly exchanged by the user (e.g., replaced by Claude).

For embedding and retrieval within the \gls{rag} process, OpenAI’s \emph{text-embedding-3-large} model and cosine similarity search were applied, with embeddings stored in a \emph{Chroma} vector store. 
The retrieval returns the top four most relevant resource interfaces based on the embedding of the natural language resource interface descriptions (chunks) and the natural language capability description as query. 
For the \emph{move-forward} capability, the following resource interfaces were returned: \texttt{cmd\_vel}, \texttt{odom}, \texttt{joint\_trajectory}, and \texttt{follow\_joint\_trajectory}. 
Among these, the two required interfaces (\texttt{cmd\_vel} and \texttt{odom}) were correctly identified. 

These selected resource interface descriptions, along with all other user inputs and the few-shot examples (capability, skill specification, and skill implementation), were then passed into the prompt for code generation.
\begin{figure}[h]
    \centering
\begin{lstlisting}[
    label=listing:skillExample,
    language=Python,
    numbers=left,
    numberstyle=\tiny\color{codegray},
    numbersep=2pt, 
    breaklines=true,
    basicstyle=\footnotesize\ttfamily,
    caption=Excerpt of \emph{move\_forward} skill generated with \emph{gpt-4o}
    ]
@skill(skill_interface=SkillInterface.REST, skill_iri="...mmo700/skills/moveForward"...)
class MoveForwardSkill(ROS2Skill): 

    def __init__(self):
        super().__init__('move_forward')
        self.velocity_publisher = None
        ...

    @skill_parameter(is_required=True, ...)
    def get_desired_distance(self) -> float:
        return self.desired_distance

    @execute
    def execute(self) -> None:
        desired_velocity = self.desired_distance / self.desired_time
        if desired_velocity > self.MAX_VELOCITY:
            ...

        self.velocity_cmd.linear.x = desired_velocity
        end_time = self.start_time + self.desired_time

        while time.time() < end_time:
            self.velocity_publisher.publish(self.velocity_cmd)
            time.sleep(0.1)

        self.velocity_cmd.linear.x = 0.0
        self.velocity_publisher.publish(self.velocity_cmd)
\end{lstlisting}
\end{figure}
Listing~\ref{listing:skillExample} shows a selected excerpt from the resulting skill implementation \emph{move\_forward} generated with \emph{gpt-4o}.
The excerpt illustrates the implementation of a class that inherits from the base class \texttt{ROS2Skill} provided by the \emph{pySkillUp} framework.
This base class includes \gls{ros2}-specific functionalities such as managing node startup and shutdown.
The \texttt{@skill} annotation is used to mark the class as a skill and includes the skill interface used for interacting with the skill as well as a generated \texttt{skill\_iri}.
The excerpt also shows an example parameter \texttt{desired\_distance} representing the target distance that the robot should travel.
Finally, the \texttt{execute} method is included, in which the required velocity is calculated based on the desired distance and duration.
If the computed velocity exceeds the maximum value, the method handles this case accordingly.
Otherwise, the velocity is published for the specified time and then reset to zero.

\subsection{Results and Discussion}
Each generated skill was evaluated in terms of structural correctness and functional behavior in a Gazebo simulation of the target mobile robot based on the following criteria: 

\begin{itemize} 
    \item \textbf{Syntax correctness} of the generated code. 
    \item \textbf{Annotation completeness:} Coverage and correctness of all required annotations, including skill metadata, parameters, outputs, and state machine states. 
    \item \textbf{Executability:} Successful execution of the generated code without runtime errors. 
    \item \textbf{Interface usage:} Correct identification of relevant resource interfaces during retrieval and their appropriate integration into the generated skill implementation. 
    \item \textbf{Behavioral accuracy:} Correct implementation of the intended capability behavior, in particular the behavior specified by the user within the state machine states.
    \item \textbf{Lines of code:} Code length as a rough indicator of implementation size and complexity.
    \item \textbf{Manual effort:} Number of modified lines required to achieve the correct behavior. 
\end{itemize}

\begin{table*}
\centering
\caption{Evaluation results for six generated skills using two \glspl{llm}: \emph{gpt-4o} and \emph{o3-mini}. Each cell shows the result as \emph{gpt-4o} / \emph{o3-mini}. Scoring: \CIRCLE = complete success; \LEFTcircle = partial success (core functionality achieved but with issues); \Circle = largely erroneous results.}
\label{tab:results}
\begin{tabularx}{\linewidth}{ l c c c c c c c c c c c }
\toprule
& Syntax & \multicolumn{4}{c}{Annotation completeness} & Executability & Interfaces & Behavior & Code lines & Manual effort \\
& & Skill & Parameters & Outputs & States &  & & & & (code lines) \\
\midrule
S1: Get Object & \CIRCLE\;/\;\CIRCLE & \CIRCLE\;/\;\CIRCLE & - & \CIRCLE\;/\;\CIRCLE & \CIRCLE\;/\;\CIRCLE & \CIRCLE\;/\;\CIRCLE & \CIRCLE\;/\;\CIRCLE & \CIRCLE\;/\;\CIRCLE & 59\;/\;82 & 0\;/\;0 \\
S2: Move Forward & \CIRCLE\;/\;\CIRCLE & \CIRCLE\;/\;\CIRCLE & \CIRCLE\;/\;\CIRCLE & \CIRCLE\;/\;\CIRCLE & \CIRCLE\;/\;\CIRCLE & \CIRCLE\;/\;\CIRCLE & \CIRCLE\;/\;\CIRCLE & \CIRCLE\;/\;\CIRCLE & 99\;/\;129 & 0\;/\;0 \\
S3: Rotate & \CIRCLE\;/\;\CIRCLE & \CIRCLE\;/\;\CIRCLE & \CIRCLE\;/\;\CIRCLE & \CIRCLE\;/\;\CIRCLE & \CIRCLE\;/\;\CIRCLE & \CIRCLE\;/\;\CIRCLE & \CIRCLE\;/\;\CIRCLE & \LEFTcircle\;/\;\CIRCLE & 95\;/\;183 & 25\;/\;0\\
S4: Collision Avoidance & \CIRCLE\;/\;\CIRCLE & \CIRCLE\;/\;\CIRCLE & \LEFTcircle\;/\;\CIRCLE & \CIRCLE\;/\;\CIRCLE & \CIRCLE\;/\;\CIRCLE & \CIRCLE\;/\;\CIRCLE & \LEFTcircle\;/\;\LEFTcircle & \CIRCLE\;/\;\LEFTcircle & 115\;/\;199 & 0\;/\;1 \\
S5: Mapping & \CIRCLE\;/\;\CIRCLE & \CIRCLE\;/\;\CIRCLE & \CIRCLE\;/\;\CIRCLE & \CIRCLE\;/\;\CIRCLE & \CIRCLE\;/\;\CIRCLE & \CIRCLE\;/\;\CIRCLE & \Circle\;/\;\LEFTcircle & \Circle\;/\;\LEFTcircle & 98\;/\;149 & 45\;/\;35 \\
S6: Move to Point & \CIRCLE\;/\;\CIRCLE & \CIRCLE\;/\;\CIRCLE & \CIRCLE\;/\;\CIRCLE & \CIRCLE\;/\;\CIRCLE & \CIRCLE\;/\;\CIRCLE & \Circle\;/\;\CIRCLE & \CIRCLE\;/\;\CIRCLE & \Circle\;/\;\LEFTcircle & 154\;/\;224 & 25\;/\;3 \\
\bottomrule
\end{tabularx}
\end{table*}

The overall results were promising, with consistently well-structured skill skeletons and most implementations exhibiting a high degree of behavioral correctness. 
A summary of the evaluation results is provided in Table~\ref{tab:results}.

Although a direct comparison between \emph{o3-mini} and \emph{gpt-4o} was not the focus of this evaluation, \emph{o3-mini} consistently demonstrated superior performance across most tasks.
It tended to produce more comprehensive code, characterized by more robust handling of edge cases and enriched with explanatory comments that improved readability and transparency.

The structural code skeleton including syntax and annotation completeness was correctly generated for all skills except \emph{collision-avoidance} generated with \emph{gpt-4o}, where one of four required parameter annotations was missing. 
Instead, the \gls{llm} inserted a hard-coded default value of $0.5$ for the minimum obstacle distance. 
This value was not mentioned in the original capability and appears to have been hallucinated.

All generated skills were executable in \gls{ros2}, with one exception: the skill \emph{move-to-point} generated with \emph{gpt-4o} failed at runtime. 
The \gls{llm} inserted placeholders instead of critical code segments, likely due to insufficient knowledge about the required logic, leading to execution failure. 
This particular skill is a complex task, and it is noteworthy that the \gls{llm} omitted rather than incorrectly implemented the unknown logic.

Regarding resource interface usage, four of six skills were implemented with the correct resource interfaces, both in retrieval and in code. 
For the \emph{collision-avoidance} skill, one of the three necessary resource interfaces for reading sensor data was not retrieved via \gls{rag}, but was nonetheless used correctly by both \glspl{llm}. 
This suggests that the \gls{llm} relied on its pretrained internal knowledge in addition to explicit retrieval in this case. 
Similarly, in the \emph{mapping} skill a motion-related resource interface was used without being retrieved, likely due to patterns observed in the few-shot examples. 
Furthermore, within the same \emph{mapping} skill generation, one additional resource interface relevant for obstacle detection was neither retrieved nor used in the implementation, resulting in missing sensor feedback handling.
A notable difference was observed between the two \glspl{llm}: while \emph{o3-mini} used the correct resource interface for accessing the map data, \emph{gpt-4o} used an unrelated resource interface, leading to an incorrect implementation of this part of the functionality.

Behavioral correctness was fully achieved in four out of six skills when considering both \glspl{llm} together.
In the case of the \emph{rotate} skill, \emph{o3-mini} produced a correct implementation, whereas \emph{gpt-4o} generated code that was functionally close but contained a critical flaw.  
Specifically, the calculation of the target rotation angle was inaccurate leading to incorrect rotation behavior. 
Fixing this issue required implementing a normalization step for the angular difference, resulting in 25 lines of modified code.

Interestingly, both models generated largely correct implementations of the complex \emph{collision-avoidance} skill.
The robot executed the desired motion pattern, and stop logic for obstacle avoidance was mostly implemented: 
\emph{o3-mini} missed only a single line that explicitly triggered the stop transition, although the obstacle detection condition was correctly implemented. 

In contrast, the two most complex skills -- \emph{mapping} and \emph{move-to-point} -- did not behave as intended. 
For \emph{mapping}, both \glspl{llm} generated basic exploration behaviors: \emph{o3-mini} produced a snake-like movement pattern, while \emph{gpt-4o} followed a simpler strategy of moving forward for the desired distance and then rotating 90 degrees.  
However, in both cases, the robot failed to avoid obstacles and collided with them. 
The corresponding feedback resource interface was neither retrieved nor implemented, likely due to an underspecified behavior description in the user input. 
In the case of \emph{o3-mini}, the missing obstacle avoidance was corrected by adding logic to adjust the robot's heading based on sensor feedback.
For \emph{gpt-4o}, more changes were necessary, including replacing incorrect map access and implementing basic avoidance strategy.
While both implementations required noticeable changes, the total number of modified lines remained manageable.

The \emph{move-to-point} skill, as previously noted, was not executable in the version generated by \emph{gpt-4o} due to some missing implementation segments and robot-specific identifiers. 
As a result, the logic for interacting with the action server remained incomplete.
In contrast, the implementation generated by \emph{o3-mini} was functionally correct, but a few robot-specific identifiers had to be manually adjusted due to the lack of such specific information in the prompt. 
After replacing these names with the correct ones, the skill executed as intended.

Overall, the evaluation shows that \textbf{LLMCap2Skill} can reliably generate executable and structurally correct skill implementations for complex tasks from capability ontologies.
While minor manual corrections were still necessary -- particularly in the implementation of behavioral logic and integration of robot-specific elements -- the core structure was consistently well-formed, and the intended functionality could in most cases be achieved with manageable effort.
Crucially, the method eliminates the need for manual skill implementation from scratch, allowing developers to shift their focus from boilerplate coding to targeted refinements.
In combination with skill frameworks such as \emph{pySkillUp}, this also includes the automatic generation of the ontology, the underlying state machine, and the skill interface required for interacting with the skill.
The combination of capabilities, \gls{rag}-based resource interface selection and code generation proved effective for translating abstract capabilities into working skills.

%% file: sections/05_Conclusion.tex
This paper introduced a novel method for generating executable skills from ontological capability descriptions using \glspl{llm} and a \gls{rag} approach. 
By combining semantic capability modeling with contextual retrieval of resource interface documentation, our method enables the automated generation of skill implementations that integrate existing libraries and frameworks (RQ1, RQ3). 
The evaluation with a mobile robot platform demonstrated that the generated skills are largely correct, executable, and significantly reduce manual coding effort (RQ2).

In the future, we plan to evaluate the method for additional frameworks beyond \gls{ros2}, including the use of \emph{SkillUp} with Java as well as PLC code, targeting alternative system interfaces and libraries. 
We also plan to enhance the retrieval step of the \gls{rag} process by incorporating richer metadata and exploring more advanced retrieval strategies.
Another promising direction is the automated refinement of skill behavior through simulation feedback or user interaction. Furthermore, integrating formal verification tools could enhance trust in the generated code and ensure safety in critical applications.